\documentclass[article, 10 pt, conference]{ieeeconf}  

\usepackage{caption}
\usepackage{subcaption}
\usepackage[dvipsnames]{xcolor}
\usepackage{graphicx}
\usepackage{ragged2e}
\usepackage[style=ieee]{biblatex}
\usepackage{amsmath, amssymb, amsfonts}
\DeclareMathOperator*{\argmin}{arg\,min}
\usepackage{float}
\usepackage{soul}
\usepackage{hyperref}
\hypersetup{
    colorlinks=true,
    linkcolor=blue,
    filecolor=blue,      
    urlcolor=blue,
    citecolor=blue,
}

\IEEEoverridecommandlockouts                              

\title{\LARGE \bf
Comprehensive Training and Evaluation on Deep Reinforcement Learning for Automated Driving in Various Simulated Driving Maneuvers
}
\author{Yongqi Dong, Tobias Datema, Vincent Wassenaar, Joris van de Weg, Cahit Tolga Kopar, and Harim Suleman
\thanks
{*This work is supported by Applied and Technical Sciences (TTW), a subdomain of the Dutch Institute for Scientific Research (NWO), Grant/Award Number: 17187}
\thanks{Yongqi Dong is with the Department of Transport and Planning, Delft University of Technology, Delft, 2628 CN, the Netherlands
        (email: \textcolor{blue}{y.dong-4@tudelft.nl}).} %
\thanks{Tobias Datema, Vincent Wassenaar, Joris van de Weg, Cahit Tolga Kopar, and Harim Suleman are with the Faculty of Electrical Engineering, Mathematics \& Computer Science, Delft University of Technology, Delft, 2628 CN, the Netherlands (email: \{\textcolor{blue}{T.Datema, V.Wassenaar, J.J.vandeWeg, C.T.Kopar, H.I.Suleman\textcolor{black}\}@student.tudelft.nl}).}%
}

\begin{document}

\maketitle
\thispagestyle{empty}
\pagestyle{empty}

\begin{abstract}

Developing and testing automated driving models in the real world might be challenging and even dangerous, while simulation can help with this, especially for challenging maneuvers. Deep reinforcement learning (DRL) has the potential to tackle complex decision-making and controlling tasks through learning and interacting with the environment, thus it is suitable for developing automated driving while not being explored in detail yet. This study carried out a comprehensive study by implementing, evaluating, and comparing the two DRL algorithms, Deep Q-networks (DQN) and Trust Region Policy Optimization (TRPO), for training automated driving on the \textit{highway-env} simulation platform. Effective and customized reward functions were developed and the implemented algorithms were evaluated in terms of on-lane accuracy (how well the car drives on the road within the lane), efficiency (how fast the car drives), safety (how likely the car is to crash into obstacles), and comfort (how much the car makes jerks, e.g., suddenly accelerates or brakes). Results show that the TRPO-based models with modified reward functions delivered the best performance in most cases. Furthermore, to train a uniform driving model that can tackle various driving maneuvers besides the specific ones, this study expanded the \textit{highway-env} and developed an extra customized training environment, namely, \textit{ComplexRoads}, integrating various driving maneuvers and multiple road scenarios together. Models trained on the designed \textit{ComplexRoads} environment can adapt well to other driving maneuvers with promising overall performance. Lastly, several functionalities were added to the \textit{highway-env} to implement this work. The codes are open on GitHub at \href{https://github.com/alaineman/drlcarsim-paper}{https://github.com/alaineman/drlcarsim-paper}. 

\end{abstract}

\section{INTRODUCTION}

Artificial intelligence (AI) is making huge improvements in various fields, one of which is automated driving \cite{badue2021self}.
One typical type of AI that is well-suitable for developing automated driving models is Deep Reinforcement Learning (DRL) \cite{rao2018deep}. DRL makes use of the advantage of deep neural networks regarding feature extraction and the advantage of reinforcement learning regarding learning from interacting with the environment. DRL exhibits excellent performance in various decision-making tasks, e.g., \textit{GO} \cite{silver2016mastering} and playing video games \cite{shao2019survey} and it has been employed in various automated driving tasks \cite{kiran2021deep,zhu2021survey,khalil2022exploiting}, e.g., lane-keeping, lane-changing, overtaking, ramp merging, and driving through intersections. 

For the lane-keeping task,  Sallab et al. \cite{sallab2016end,sallab2017deep} developed DRL-based methods for delivering both discrete policies using Deep Q-network (DQN) and continuous policies using Deep Deterministic Actor-Critic Algorithm (DDAC) to follow the lane and to maximize the average velocity when driving on the curved race track on Open Racing Car Simulator (TORCS). Similarly, for the lane-changing task, Wang et al. \cite{wang2018reinforcement} trained a DQN-based model to perform decision-making of lane-keeping, lane changing to the left/right, and acceleration/deceleration, so that the trained agent can intelligently make a lane change under diverse and even unforeseen scenarios. Furthermore, Zhang et al. \cite{zhang2022multi} proposed a bi-level lane-change behavior planning strategy using DRL-based lane-change decision-making model and negotiation-based right-of-way assignment model to deliver multi-agent lane-change maneuvers. For the overtaking task, Kaushik et al. \cite{kaushik2018overtaking} adopted Deep Deterministic Policy Gradients (DDPG) to learn overtaking maneuvers for an automated vehicle in the presence of multiple surrounding cars in a simulated highway scenario. They verified that their curriculum learning resembled approach can learn to smooth overtaking maneuvers, largely collision-free, and independent of the track and number of cars in the scene. For the ramp merging task, Wang and Chan \cite{wang2017formulation} employed a Long-Short Term Memory (LSTM) neural network to model the interactive environment conveying internal states containing historical driving information to a DQN which then generated Q-values for action selection regarding on-ramp merging. Additionally, for negotiating and driving through intersections, Isele et al. \cite{isele2018navigating} explored the effectiveness of the DQN-based DRL method to handle the task of navigating through unsignaled intersections. Finally, Guo and Ma \cite{guo2021drl} developed a real-time learning and control framework for signalized intersection management, which integrated both vehicle trajectory control and signal optimization using DDPG-based DRL learning directly from the dynamic interactions between vehicles, traffic signal control and traffic environment in the mixed connected and automated vehicle (CAV) environment.

\begin{figure}[!ht]
\vspace{-0.95em}
    \centering
    \includegraphics[width=0.33\textwidth]{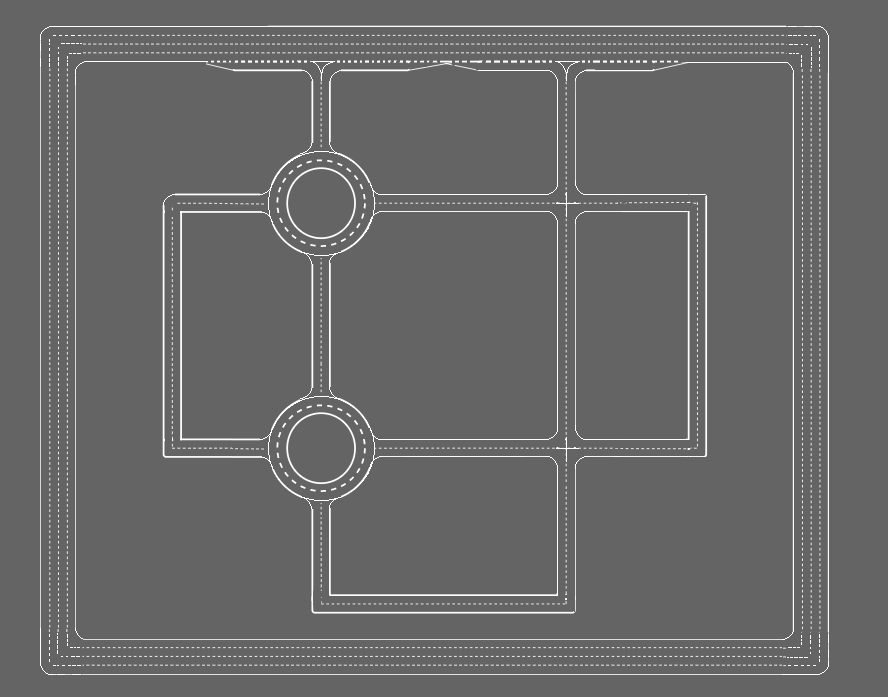}
    \caption{The layout of \textit{ComplexRoads} environment}
    \label{fig:complex}
\vspace{-0.95em}  
\end{figure}

It is observed that although many studies have utilized DRL for various driving tasks, most of them focus only on one specific driving maneuver. Seldom do they evaluate the DRL model performance across different maneuvers and neither do they explore the adaptability of DRL models trained on one specific environment but tested in other various maneuvers. This study tries to fill this research gap by implementing, evaluating, and comprehensively comparing the performance of two DRLs, i.e., DQN and TRPO, in various driving scenarios.  Customized effective reward functions were developed and the implemented DRLs were evaluated in terms of various aspects considering driving safety, efficiency, and comfort level. This study also constructed a new simulation environment, named \textit{`ComplexRoads'} (shown in Fig \ref{fig:complex}), integrating various driving maneuvers and multiple road scenarios. The \textit{ComplexRoads} served to train a uniform driving model that can tackle various driving tasks. And to verify this, the models trained only on \textit{ComplexRoads} were tested and evaluated in the specific driving maneuvers. Intensive experimental results demonstrated the effectiveness of this customized training environment.

To advance the learning capability for the developed DRL-based AI models, i.e. encouraging relational insight, besides designing \textit{ComplexRoads}, several built-in functions of the \texttt{highway-env} package were also upgraded. Notable modifications are summarized as follows: the tracking of the `current' lane with respect to the car (training agent) was upgraded to take into account the lane heading to eliminate confusing transitions when driving off-road. Furthermore, the distance between the car and its current lane was upgraded to a signed value to allow for orientation distinction. Similarly, the lane heading difference, LHD for short, was adjusted to also be a signed value. These improvements yield increased learning abilities for both on-road driving, returning to on-road driving when off-road, and a general sense of `awareness' given an arbitrary environment.


\section{METHODOLOGY}

\subsection{System Framework}
\begin{figure}[!ht]
    \centering
    \includegraphics[width=0.5\textwidth]{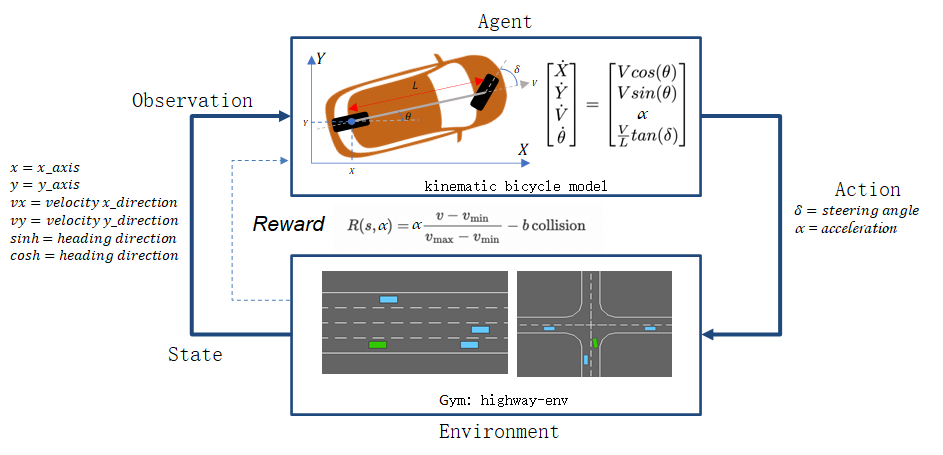}
    \caption{The system framework-illustration of the DRL MDP.}
    \label{fig:learningcycle}
\vspace{-1.95em}   
\end{figure}

The general DRL learning cycle is an iterative learning process based on the agent's performance in the environment influenced by the agent's actions. In mathematical terms, automated driving can be modeled as a Markov Decision Process (MDP) \cite{hu2007markov}. MDP captures the features of sequential decision-making. The components of an MDP include environments, agents, actions, rewards, and states. In this study, the system framework which illustrates the corresponding MDP is depicted in Fig \ref{fig:learningcycle}. The system generally consists of five main elements, i.e., environment, agent, action, state, and reward, which will be elaborated in detail in this section.


\subsection{DRL MDP Elements}


\textit{Environment}: To simulate the MDP, this study adopted the \textit{highway-env} platform \cite{highway-env}, which is a Python-based package that offers a variety of driving environments. 
As a widely used platform, ample research has been conducted using the \textit{highway-env}, such as \cite{alizadeh2019automated,liu2022autonomous}.
In the \textit{highway-env}, six dedicated driving scenarios are available, i.e., Highway, Merge, Roundabout, Intersection, Racetrack, and Parking. Users can also customize environments by specifying the number of lanes, the size of a given roundabout, and other parameters. In this study, all the driving scenarios, except for the Highway and Parking,  are covered.

For training and evaluating a uniform driving model, this study designed a new simulation environment, named \textit{`ComplexRoads'} (shown in Fig \ref{fig:complex}). \textit{`ComplexRoads'} integrates two highway merging scenarios, two four-way intersections, two roundabouts, and several segments of multi-straight lanes. The DRL models trained only on \textit{ComplexRoads} were tested and evaluated in the specific driving maneuvers originally available on \textit{highway-env}.

\textit{Agent}: A kinematic bicycle model is used to represent the vehicle as the agent of MDP. Despite its simplicity, a kinematic bicycle model is able to represent actual vehicle dynamics \cite{polack2017kinematic}.

\textit{Action}: An action taken by the agent in the proposed MDP is an element from the contracted \textit{Action Space}. In this study,  the two dimensions of the Action Space $\mathcal{A}$ are: acceleration (throttle) and steering angle of the front wheels. Depending on the DRL algorithm $\mathcal{A}$ is either of the form $\left[-\frac{\pi}{2},\frac{\pi}{2}\right]\times[-5,5]$ for algorithms requiring a continuous action space, or $\{\delta_1,\ldots,\delta_n\} \times \{\alpha_1,\ldots\,\alpha_m\}$ in the $n \times m$ discrete case. Hence, $(\delta,\alpha) \in \mathcal{A}$, where steering is denoted by $\delta$ and acceleration is denoted by $\alpha$.

\textit{State}: As illustrated in Fig \ref{fig:learningcycle}, the state in the proposed MDP includes the ego AV’s state, e.g., location $(x,y)$, velocity $(v_x,v_y)$, and heading direction, together with the surrounding vehicles state and road conditions and is directly accessible at each time frame to the ego car, either in absolute terms or relative to itself.

\textit{Reward}: The customized \textit{Reward} function is elaborated in detail in the following subsection \textit{C}.


\subsection{Reward Function}
For training the models, this study used the reward function already present in the \textit{highway-env} package (referred to as the baseline reward and is illustrated in the middle of Fig \ref{fig:learningcycle}) and the own modified and upgraded reward function. The model performances were compared to demonstrate that the upgraded reward is better than the baseline reward. During the training it is observed that in the early stages, the trained agent car would sometimes drive off the road. To make the training more efficient in handling the off-road driving and stimulating the agent to return to driving on-road, one specific contribution in this study is to adjust the distance measure between the agent and the lane, in addition with constructing the lane heading difference measure illustrated in the following paragraphs.

Let $c$ denote the ego car agent and $\mathcal{L}$ the corresponding lane. A lane is a collection of lane points $l\in \mathcal{L}$. Now define $l'$ as the lane point with the shortest Euclidean distance to the car, meaning
\begin{equation}
    l' := \argmin_{l\in \mathcal{L}} d(c,l')
\end{equation}
and define the orientation $\omega$ of the car $c$ with respect to a lane point $l$ as follows
\begin{equation}
    \omega(c,l)= \begin{cases}
        1 & \text{if car is located left of $l$\footnotemark} \\
        -1 & \text{otherwise}
    \end{cases}    
\end{equation}
\footnotetext{More precisely; if the car is located left of the tangent line for the lane segment containing $l$}
Then, this study defines the distance between the ego car and the lane as the shortest distance from the ego car $c$, to any point $l$ on lane $\mathcal{L}$, meaning
\begin{equation}
    d(c,\mathcal{L}) = \omega(c,l') d(c,l')
\end{equation}
The car heading and lane point heading are denoted by $c_\varphi$ and $l_\varphi$ respectively, both values are within angle range $(-\pi, \pi]$. Now, the lane heading difference (LHD) is defined as
\begin{equation}
    \mathrm{LHD} = \begin{cases}
        l_\varphi - c_\varphi + 2\pi& \text{if } l_\varphi - c_\varphi < -\pi \\
        l_\varphi - c_\varphi - 2\pi & \text{if } l_\varphi - c_\varphi > \pi\\
        l_\varphi - c_\varphi & \text{otherwise}
    \end{cases}
\end{equation}
An important remark to this setup is the fact that if $\text{sgn}(\mathrm{LHD}) \cdot \text{sgn}(d(c,\mathcal{L}))<0$ then the car is heading for the lane. Similarly, if $\text{sgn}(\mathrm{LHD}) \cdot \text{sgn}(d(c,\mathcal{L}))>0$ the car is deviating (further) from the lane.
\begin{figure}[!ht]
     \centering
     \begin{subfigure}[b]{0.2\textwidth}
         \centering
         \includegraphics[width=0.9\textwidth]{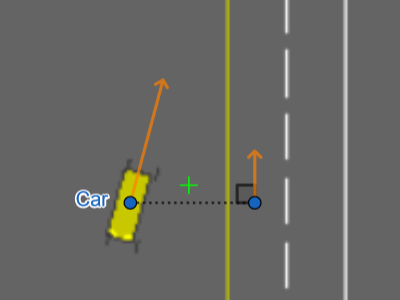}
        \caption{Off-road scenario with $d(c,l')>0$ and $\mathrm{LHD}<0$.}
         \label{fig:dpos-lhdneg}
     \end{subfigure}\hspace{7mm}     
     \begin{subfigure}[b]{0.2\textwidth}
        \centering
         \includegraphics[width=0.9\textwidth]{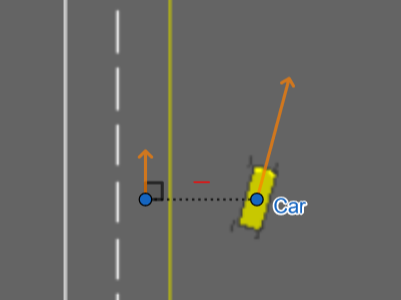}
         \caption{Off-road scenario with $d(c,l')<0$ and $\mathrm{LHD}<0$.}
         \label{fig:dneg-lhdneg}
     \end{subfigure}\vspace{0.75em} 
     \begin{subfigure}[b]{0.2\textwidth}
         {\centering
         \includegraphics[width=0.9\textwidth]{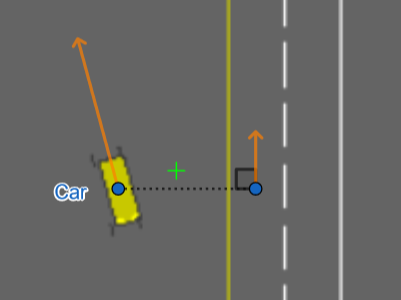}
         \caption{Off-road scenario with $d(c,l')>0$ and $\mathrm{LHD}>0$.}
         \label{fig:dpos-lhdpos}}
     \end{subfigure} \hspace{7mm}
    \begin{subfigure}[b]{0.20\textwidth}        
        \centering
          \includegraphics[width=0.9\textwidth]{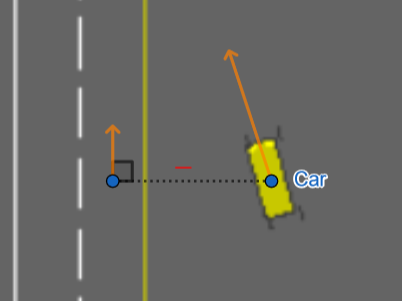}
           \caption{Off-road scenario with $d(c,l')<0$ and $\mathrm{LHD}>0$.}
         \label{fig:dneg-lhdpos}
     \end{subfigure}
    \caption{Four different off-road scenarios showcasing available environment observations of the ego car. Both lane heading and car heading are portrayed by vectors. The lane distance and $\mathrm{LHD}$, for the ego car $c$ with respect to the lane point $l'$. The sign is orientation based: if the car is located left of the road, the Euclidean distance is perceived as positive, and negative if located right of the road.}
    \label{fig: four graphs}
    \vspace{-1.75em} 
\end{figure}

Finally, denote the velocity of the ego car $c$ by $c_v$, the reward function $R:\mathbb{R}^3 \rightarrow \mathbb{R}$, with regard to \textit{state} $S$, is defined as
\begin{equation}
    \label{eq:Reward}
    R_S(c,\mathcal{L})=
    \begin{cases}
        \frac{\cos(\lvert \mathrm{LHD} \rvert)\cdot c_v}{20\cdot\max(1,|d(c,\mathcal{L}|)} & \text{if } c_v\geq 0\\
        0 & \text{otherwise}
    \end{cases}
\end{equation}
where $\mathrm{LHD}$ is the lane heading difference between the ego car and the closest lane point. However, if the car crashes during the simulation, the reward is automatically set as -10, regardless of the \textit{state}. 

The reward function, as defined in Equation \ref{eq:Reward}, rewards the car for its `effective' speed on the road, defined by the cosine of the angular difference between the direction the car is driving in and the direction in which the road goes, multiplied by the speed of the car. With this design, both an increase in the driving speed and driving in line with the road heading will result in high rewards. Moreover, the value is divided by the lane offset to punish the car for driving off-road and also divided by 20 to scale the reward function to remain close to 1 under optimal circumstances.

\subsection{DRL Algorithms}
Regarding DRL algorithms, TRPO  \cite{schulman2015trust} and DQN  \cite{fan2020theoretical} were customized and implemented. Details of the DRLs including hyperparameter settings are elaborated in the supplementary at \href{https://drive.google.com/drive/folders/1IrXCxATJucIpF3RUARRVpJfQUITcJ3z8}{https://shorturl.at/oLP57}, while Section \ref{sec:Results} presents the results comparing trained DRLs' performances.

\subsection{Evaluation of the Models}
To evaluate and compare the model performance, one needs a set of indicators and metrics. For which this study implemented a performance logger that measures and stores various indicators when testing a model in a given environment. These indicators are measured for a set amount of runs and the logger then prints the average values over all the runs. The measured indicators are: 1) Speed, 2) Peak jerk, 3) Total jerk, 4) Total distance, 5) Total steering, 6) Running time, 7) Lane time (rate of time the car is running within the road), and 8) Rate of collision.



The jerk is defined as the difference between the current and the previous action of a vehicle, consisting of both the steering angle and the acceleration. 
The magnitude of the total jerk reflects the degree to which the vehicle's motion changes abruptly and frequently, where a higher value of the total jerk implies a less comfortable driving.  The jerk is defined by equations in \ref{eqn:jerk}:
\vspace{-0.35em} 
\begin{align}
\begin{split}
    \label{eqn:jerk}
    J_\text{acceleration} = \frac{a_{t-1} - a_{t}}{a_{\max} - a_{\min}}\\
    J_\text{steering} = \frac{w_{t} - w_{t-1}}{w_{\max} - w_{\min}}\\
        J_{\text{total}} = \frac{J_\text{acceleration} + J_\text{steering}}{2} 
\end{split}
\end{align}

The total steering is defined as the total sum of steering the car performs in the course of an evaluation, measured in angles. A higher amount of steering could, to certain extent, imply less efficient driving with unnecessary steering.

The onlane rate is defined as the amount of time the evaluated car spends driving on the lane, divided by the total amount of time the car spends driving. The collision rate is defined as the total amount of collisions the car makes, divided by the total amount of evaluation trials.

\section{Experiments}

This study conducted intensive experiments to train and evaluate DRL models using TRPO and DQN algorithms on four environments provided by \textit{highway-env}, and also the newly self-designed \textit{ComplexRoads}. The models were trained using both the original standard reward function provided by \textit{highway-env} (which served as the baseline) and the customized reward function. The hyper-parameters used for training can be found in the appendix at \href{https://drive.google.com/drive/folders/1IrXCxATJucIpF3RUARRVpJfQUITcJ3z8}{https://shorturl.at/oLP57}. The models were trained on the supercomputer Delft Blue \cite{DHPC2022}. For every environment, ten models were trained and saved for 10,000 and 100,000 iterations. When finishing training, the model performance was tested for 10 runs. During the performance testing, constraints such as a maximum running time, minimum speed and if a crash had occurred were adopted. To obtain an overall assessment, the average of all these 10 testing results was calculated. To get an idea of how well the models perform regarding an uniform driving model, they were not only tested in their trained environments, but also cross-evaluated in other different environments. With the cross-evaluating, the effectiveness of the newly designed environment \textit{ComplexRoads} can be verified. The experiment testing results are summarized and discussed in Section \ref{sec:Results}. 

\section{Results and discussion}
\label{sec:Results}


Tables \ref{t:comp}, \ref{t:round}, \ref{t:int}, \ref{t:merge}, and \ref{t:race} present the average performances of the DRL models trained on five environments and evaluated on the same respective environment. For every model variant in one specific environment, this study trained it for 10 times and also evaluated it for 10 times to get the average performance indicators. This paper writes ``1*" when the number is rounded to 1, but not quite equal to 1. With the letters ``B'' and ``M'', this paper refers to whether the baseline reward function or modified reward function was used in training the model.

Meanwhile, Tables \ref{t:complex-DQN-cross}, \ref{t:complex-TRPO-cross}, \ref{t:round-DQN-cross}, \ref{t:inters-DQN-cross}, \ref{t:race-DQN-cross}, and \ref{t:merge-DQN-cross} present the average performances of the implemented DRL models trained in their own environment, but evaluated in other different environments. This is for evaluating how adaptive these models are. In order to save space, these tables leave out some of the `less important' indicators, which can be found in the appendix at \href{https://drive.google.com/drive/folders/1IrXCxATJucIpF3RUARRVpJfQUITcJ3z8}{https://shorturl.at/oLP57}. 

One needs to note that for the environment of Merge and the self-designed \textit{ComplexRoads}, no baseline reward functions are available, so only the models trained by the modified and upgraded reward (indicated with ``-M") were evaluated. Also, for cross-environments evaluating, only models with the modified reward were evaluated.

\begin{table}[H]
\vspace{-0.1em}
\begin{center}
\caption{\textit{ComplexRoads}}
\label{t:comp}
\begin{tabular}{ | c | c | c |}
 \hline
 \textbf{Indicator}  & \textbf{DQN-M} & \textbf{TRPO-M}\\ 
 \hline
 speed  & 16.1  & 16.2 \\
 \hline  
 pk. jerk  & 0.99  & 0.799 \\
 \hline
 tot. jerk  & 221  & 13.2\\
 \hline
 tot. distance  & 661  & 547\\
 \hline
 tot. steering  & 263  & 46.3\\
 \hline
 runtime  & 607  & 492\\
 \hline
 onlane rate  & 0.999  & 0.999\\
 \hline
 col. rate  & 0.07  & 0.09\\
 \hline
\end{tabular}
\end{center}
\vspace{-2.0em}
\end{table}

\begin{table}[H]
\vspace{-0.6em}
\begin{center}
\caption{Roundabout}
\label{t:round}
\begin{tabular}{ | c | c | c | c | c |}
 \hline
 \textbf{Indicator} & \textbf{DQN-B} & \textbf{DQN-M} & \textbf{TRPO-B} & \textbf{TRPO-M}\\ 
 \hline
 speed & 8.3 & 8.6 & 7.88 & 8.25 \\
 \hline
 pk. jerk & 1.07 & 1.09 & 0.704 & 0.775 \\
 \hline
 tot. jerk & 71 & 159 & 15.4 & 20.7\\
 \hline
 tot. distance & 185 & 278 & 214 & 229\\
 \hline
 tot. steering & 128 & 210 & 92.8 & 114\\
 \hline
 runtime & 318 & 479 & 382 & 394\\
 \hline
 onlane rate & 0.384 & 0.783 & 0.341 & 0.693\\
 \hline
 col. rate & 0.71 & 0.68 & 0.51 & 0.62\\
 \hline
\end{tabular}
\end{center}
\vspace{-2.0em}
\end{table}

\begin{table}[H]
\begin{center}
\vspace{-0.7em}
\caption{Intersection}
\label{t:int}
\begin{tabular}{ | c | c | c | c | c |}
 \hline
 \textbf{Indicator} & \textbf{DQN-B} & \textbf{DQN-M} & \textbf{TRPO-B} & \textbf{TRPO-M}\\ 
 \hline
 speed & 9.89 & 10.1 & 9.74 & 10.3 \\
 \hline  
 pk. jerk & 0.892 & 1.04 & 0.545 & 0.637 \\
 \hline
 tot. jerk & 24.3 & 32.6 & 6.21 & 6.53\\
 \hline
 tot. distance & 38.6 & 62.8 & 65 & 68.3\\
 \hline
 tot. steering & 29.9 & 41.4 & 18.7 & 18.5\\
 \hline
 runtime & 58.8 & 93.3 & 101 & 100\\
 \hline
 onlane rate & 0.988 & 0.999 & 0.999 & 1*\\
 \hline
 col. rate & 0.38 & 0.49 & 0.33 & 0.19\\
 \hline
\end{tabular}
\end{center}
\vspace{-1.5em}
\end{table}

\begin{table}[H]
\vspace{-0.8em}
\begin{center}
\caption{Merge}
\label{t:merge}
\begin{tabular}{ | c | c | c |}
 \hline
 \textbf{Indicator}  & \textbf{DQN-M} & \textbf{TRPO-M}\\ 
 \hline
 speed  & 30.9 & 29.1 \\
 \hline  
 pk. jerk  & 0.863  & 0.607 \\
 \hline
 tot. jerk  & 47.8  & 11.2\\
 \hline
 tot. distance  & 491 & 487\\
 \hline
 tot. steering  & 86.7  & 82.1\\
 \hline
 runtime  & 226 & 253\\
 \hline
 onlane rate  & 0.875 & 0.836\\
 \hline
 col. rate  & 0.5  & 0.4\\
 \hline
\end{tabular}
\end{center}
\vspace{-2.0em}
\end{table}

\begin{table}[H]
\vspace{-0.2em}
\begin{center}
\caption{Racetrack}
\label{t:race}
\begin{tabular}{ | c | c | c | c | c |}
 \hline
 \textbf{Indicator} & \textbf{DQN-B} & \textbf{DQN-M} & \textbf{TRPO-B} & \textbf{TRPO-M}\\  \hline
 speed & 7.12 & 9.44 & 10.3 & 7.59 \\
 \hline  
 pk. jerk & 0.956 & 0.756 & 0.518 & 0.962 \\
 \hline
 tot. jerk & 70.6 & 43.1 & 8.35 & 127\\
 \hline
 tot. distance & 229 & 207 & 254 & 222\\
 \hline
 tot. steering & 183 & 67.5 & 84.5 & 181\\
 \hline
 runtime & 449 & 346 & 362 & 471\\
 \hline
 onlane rate & 0.225 & 0.991 & 0.943 & 0.992\\
 \hline
 col. rate & 0.13 & 0.84 & 0.74 & 0.29\\
 \hline
\end{tabular}
\end{center}
\end{table}

\begin{table}[H]
\vspace{-0.1em}
\begin{center}
\caption{DQN-M trained on \textit{ComplexRoads} evaluated in other various environments}
\label{t:complex-DQN-cross}
\begin{tabular}{ | c | c | c | c | c |}
 \hline
 \textbf{Indicator} & \textbf{Racetrack} & \textbf{Roundabout} & \textbf{Merge} & \textbf{Intersection}\\ 
 \hline
 speed & 10.2 & 8.3 & 30.6 & 10 \\
 \hline  
 tot. distance & 180 & 200 & 377 & 59.3 \\
 \hline  
 runtime & 275 & 349 & 185 & 89.5 \\
 \hline  
 onlane rate & 0.998 & 0.602 & 0.935 & 0.998 \\
 \hline 
 col. rate & 0.92 & 0.79 & 0.3 & 0.52 \\  
 \hline
\end{tabular}
\end{center}
\end{table}

\begin{table}[H]
\vspace{-2.3em}
\begin{center}
\caption{TRPO-M trained on \textit{ComplexRoads} evaluated in other various environments}
\label{t:complex-TRPO-cross}
\begin{tabular}{ | c | c | c | c | c |}
 \hline
 \textbf{Indicator} & \textbf{Racetrack} & \textbf{Roundabout} & \textbf{Merge} & \textbf{Intersection}\\ 
 \hline
 speed & 9.99 & 8.95 & 29.8 & 10.3 \\
 \hline  
 tot. distance & 130 & 195 & 339 & 59.7 \\
 \hline  
 runtime & 222 & 289 & 172 & 87.3 \\  
 \hline
 onlane rate & 1* & 0.647 & 0.996 & 0.999 \\  
 \hline
 col. rate & 0.82 & 0.76 & 0.1 & 0.51 \\  
 \hline
\end{tabular}
\end{center}
\end{table}

\begin{table}[H]
\vspace{-2.3em}
\begin{center}
\caption{DQN-M trained on Roundabout evaluated in other various environments}
\label{t:round-DQN-cross}
\begin{tabular}{ | c | c | c | c |}
 \hline
 \textbf{Indicator} & \textbf{Racetrack} & \textbf{Merge} & \textbf{Intersection}\\ 
 \hline
 speed & 10.7 & 30.6 & 10.1 \\  
 \hline
 tot. distance & 156 & 335 & 22.3 \\  
 \hline
 runtime & 224 & 164 & 32.6 \\  
 \hline
 onlane rate & 0.954 & 0.955 & 0.968 \\  
 \hline
 col. rate & 0.97 & 0.2 & 0.05 \\  
 \hline
\end{tabular}
\end{center}
\end{table}

\begin{table}[H]
\vspace{-2.3em}
\begin{center}
\caption{TRPO-M trained on Intersection evaluated in other various environments}
\label{t:inters-DQN-cross}
\begin{tabular}{ | c | c | c | c |}
 \hline
 \textbf{Indicator} & \textbf{Racetrack} & \textbf{Merge} & \textbf{Roundabout}\\ 
 \hline
 speed & 9 & 30.9 & 8.91 \\  
 \hline
 tot. distance & 137 & 477 & 236 \\  
 \hline
 runtime & 253 & 228 & 345 \\  
 \hline
 onlane rate & 0.999 & 0.97 & 0.527 \\  
 \hline
 col. rate & 0.57 & 0.1 & 0.68 \\  
 \hline
\end{tabular}
\end{center}
\end{table}

\begin{table}[H]
\vspace{-2.3em}
\begin{center}
\caption{TRPO-M trained on Merge evaluated in other various environments}
\label{t:merge-DQN-cross}
\begin{tabular}{ | c | c | c | c |}
 \hline
  \textbf{Indicator} & \textbf{Intersection}  & \textbf{Racetrack} & \textbf{Roundabout}\\ 
 \hline
 speed & 9.87 & 9.85 & 9.38 \\  
 \hline
 tot. distance & 14.2 & 437 & 349 \\  
 \hline
 runtime & 22.2 & 632 & 486 \\  
 \hline
 onlane rate & 0.886 & 0.399 & 0.159 \\  
 \hline
 col. rate & 0.06 & 0.16 & 0.38 \\  
 \hline
\end{tabular}
\end{center}
\end{table}

\begin{table}[H]
\vspace{-2.3em}
\begin{center}
\caption{TRPO-M trained on Racetrack evaluated in other various environments}
\label{t:race-DQN-cross}
\begin{tabular}{ | c | c | c | c |}
 \hline
  \textbf{Indicator} & \textbf{Intersection}  & \textbf{Merge} & \textbf{Roundabout}\\ 
 \hline
 speed & 9.69 & 29.7 & 7.38 \\  
 \hline
 tot. distance & 50.9 & 304 & 113 \\  
 \hline
 runtime & 79.3 & 154 & 239 \\  
 \hline
 onlane rate & 0.996 & 0.971 & 0.849 \\  
 \hline
 col. rate & 0.67 & 0.6 & 0.76 \\  
 \hline
\end{tabular}
\end{center}
\end{table}
\vspace{-1.3em}
While there might be  various ways to express that one model outperforms another, it is important to prioritize safety as the main concern. Therefore, the measured values that this study considers the most important are the onlane rate and the collision rate, which reflect driving safety. Other values, such as speed or jerk, are less important but can be compared in cases where the onlane and collision rates are similar.

From Tables \ref{t:comp}, \ref{t:round}, \ref{t:int}, \ref{t:merge} and \ref{t:race}, one can see that in most cases the DQN with modified reward function (DQN-M) and the TRPO with modified reward function (TRPO-M) outperform the DQN and TRPO models with the baseline reward functions, especially with regards to the onlane rate.

Between the DQN and TRPO models, the models trained by TRPO tend to perform somewhat better in most cases.

Furthermore, looking at Tables \ref{t:complex-DQN-cross}, \ref{t:complex-TRPO-cross}, \ref{t:round-DQN-cross}, \ref{t:inters-DQN-cross}, \ref{t:race-DQN-cross} and \ref{t:merge-DQN-cross}, it is observed that the models trained on \textit{ComplexRoads} indeed tend to perform better than the other models in the cross-evaluation, especially in keeping a high onlane rate. This is due to various traffic situations represented in the \textit{ComplexRoads} environment, as well as the fact that the starting location of the car during training on \textit{ComplexRoads} was randomized, meaning that the car can experience various driving situations. This will also prevent the model from merely `memorizing' the environment, but instead learning better to master the maneuvers to interact with the randomly generated environments.

Due to the size of \textit{ComplexRoads}, training on it was very computationally intensive, especially with a large amount of simulated surrounding cars. Non-ego cars get destinations assigned randomly and drive around scripted, meaning they follow deterministic driving rules to drive `perfectly' and receive a new destination upon reaching the previous one. Thus, this study opted to train the model with a relatively few surrounding cars, meaning that the model does not get to interact with other cars as often as in the other environments. Due to this, it resulted in a higher collision rate when evaluated in the other environments with more surrounding cars. When the computational resource is abundant, by adding more surrounding cars into the \textit{ComplexRoads} environment, this reduced awareness of the ego car can be reduced.

All in all, it is verified that the designed \textit{ComplexRoads} indeed contributes to the training of a more flexible and adaptive driving model. All the testing scenarios and results are better demonstrated in the appendix with the demo videos also provided at 
\href{https://drive.google.com/drive/folders/1IrXCxATJucIpF3RUARRVpJfQUITcJ3z8}{https://shorturl.at/oLP57}.

\section{CONCLUSION}
This study first summarized the utilization of DRL in every specific automated driving task, e.g., lane-keeping, lane-changing, overtaking, and ramp merging, then customized and implemented two widely used DRLs, i.e., DQN and TRPO to tackle various driving maneuvers and carried out a comprehensive evaluation and comparison on the model performance. Based on \texttt{highway-env}, a modified and upgraded reward function was designed for training the DRL models. Furthermore, a new integrated training environment, \textit{ComplexRoads}, was constructed, together with several built-in functions were upgraded. 
Through various experiments, it is verified that the models trained using the modified reward generally outperformed those with the original baseline reward and the newly constructed \textit{ComplexRoads} demonstrated effective performance in training a uniform model that can tackle various driving tasks rather than one specific maneuver. As a preliminary study, the findings will provide meaningful and instructive insights for future studies towards developing automated driving with DRL and simulation.

 One feature that was implemented by this study but was removed due to time constraints and lack of computational resources, was training the cars to reach a specific destination in the designed \textit{ComplexRoads} environment, which requires more interactions of training and perhaps the implementation of a path finding and optimization algorithm. In particular, providing a metric distance from the ego car to the destination, incentivized the car to take road options which seem to directly reduce the distance, which means the car often choose poorly and got punished by the distance increasing before decreasing. While this approach might work for grid like city structures it confuses the learning process in general. Nevertheless alternative direct navigation reward designs are a very interesting direction for further research. 

\addtolength{\textheight}{-20cm}   









\printbibliography
\end{document}